\documentclass[lettersize,journal]{IEEEtran}
\usepackage{amsmath,amsfonts}
\usepackage{array}

\usepackage[caption=false, font=footnotesize, labelformat=empty]{subfig}
\usepackage{textcomp}
\usepackage{stfloats}
\usepackage{url}
\usepackage{verbatim}
\usepackage{graphicx}
\usepackage{cite}
\usepackage{amssymb} 

\usepackage[table]{xcolor}
\definecolor{Viri1Base}{HTML}{440154}
\definecolor{Viri2Base}{HTML}{3B528B}
\definecolor{Viri3Base}{HTML}{21918C}
\definecolor{Viri4Base}{HTML}{5EC962}
\definecolor{Viri5Base}{HTML}{FDE725}

\colorlet{colour1}{Viri1Base!25!white}
\colorlet{colour2}{Viri2Base!25!white}
\colorlet{colour3}{Viri3Base!25!white}
\colorlet{colour4}{Viri4Base!25!white}
\colorlet{colour5}{Viri5Base!25!white}

\usepackage{stackengine}

\usepackage[ruled,vlined]{algorithm2e}

\usepackage{stmaryrd}

\newcolumntype{C}[1]{>{\centering\arraybackslash}m{#1}}
\newcolumntype{R}[1]{>{\raggedright\arraybackslash}m{#1}}
\newcolumntype{L}[1]{>{\raggedleft\arraybackslash}m{#1}}

\makeatletter
\def\bstctlcite{\@ifnextchar[{\@bstctlcite}{\@bstctlcite[@auxout]}}
\def\@bstctlcite[#1]#2{\@bsphack
  \@for\@citeb:=#2\do{%
    \edef\@citeb{\expandafter\@firstofone\@citeb}%
    \if@filesw\immediate\write\csname #1\endcsname{\string\citation{\@citeb}}\fi}%
  \@esphack}
\makeatother

\usepackage{booktabs}
\usepackage{multirow}

\usepackage[T1]{fontenc}

\usepackage{calc} 

\usepackage[pdfstartview=XYZ,
bookmarks=true,
colorlinks=true,
linkcolor=blue,
urlcolor=blue,
citecolor=blue,
bookmarks=true,
linktocpage=true, 
hyperindex=true
]{hyperref}

\usepackage{orcidlink}

\begin{document}
\bstctlcite{IEEEexample:BSTcontrol} 

\title{Imitation Game for Adversarial Disillusion with Chain-of-Thought Reasoning in Generative AI}

\author{Ching-Chun Chang, Fan-Yun Chen, Shih-Hong Gu, Kai Gao, Hanrui Wang, and Isao Echizen

\thanks{C.-C. Chang, H. Wang and I. Echizen are with the Information and Society Research Division, National Institute of Informatics, Tokyo, Japan}
\thanks{F.-Y. Chen, S.-H. Gu and K. Gao are with the Department of Information Engineering and Computer Science, Feng Chia University, Taichung, Taiwan.}
\thanks{Correspondence: C.-C. Chang (email: ccchang@nii.ac.jp)}
}

\maketitle

\begin{abstract}
As the cornerstone of artificial intelligence, machine perception confronts a fundamental threat posed by adversarial illusions. These adversarial attacks manifest in two primary forms: deductive illusion, where specific stimuli are crafted based on the victim model's general decision logic, and inductive illusion, where the victim model's general decision logic is shaped by specific stimuli. The former exploits the model's decision boundaries to create a stimulus that, when applied, interferes with its decision-making process. The latter reinforces a conditioned reflex in the model, embedding a backdoor during its learning phase that, when triggered by a stimulus, causes aberrant behaviours. The multifaceted nature of adversarial illusions calls for a unified defence framework, addressing vulnerabilities across various forms of attack. In this study, we propose a disillusion paradigm based on the concept of an imitation game. At the heart of the imitation game lies a multimodal generative agent, steered by chain-of-thought reasoning, which observes, internalises and reconstructs the semantic essence of a sample, liberated from the classic pursuit of reversing the sample to its original state. As a proof of concept, we conduct experimental simulations using a multimodal generative dialogue agent and evaluates the methodology under a variety of attack scenarios. Experimental results demonstrate that the proposed framework consistently neutralises both deductive and inductive adversarial illusions across diverse white-box and black-box attack scenarios.
\end{abstract}

\section{Introduction}
\IEEEPARstart{M}{achine} perception refers to the capability of a computational system to process data in a way that mimics how humans use their senses to interpret and interact with the world around them~\cite{Turing:1950aa, Rosenblatt:1958aa, LeCun:2015aa}. Adversarial illusions pose an insidious threat to the safety of machine perception, ranging from classic pattern recognition tasks such as object recognition in robotics and vehicular automation to modern cybersecurity applications such as malware detection and synthetic content identification in surveillance and forensic systems~\cite{10.1145/1014052.1014066, 10.1145/1081870.1081950, 10.1145/2046684.2046692, Biggio:2018aa}. Such adversarial threats have led to catastrophic consequences in economy, politics and society, including financial fraud, election manipulation and the erosion of public trust in digital media. These attacks typically manifest in two distinct forms. One is the exploitation of the vulnerabilities in a model’s decision boundaries to craft an illusory stimulus that appears imperceptible to humans yet elicits deviant decisions from the model~\cite{DBLP:journals/corr/SzegedyZSBEGF13, 43405, Kurakin:2017aa, 8099500, 10.1145/3134599, Madry:2018aa, 8953423, 8601309, NEURIPS2019_e2c420d9, pmlr-v119-croce20b}. Another is the poisoning of learning data to implant a backdoor within a model, conditioning it towards erroneous behaviours when triggered by an illusory stimulus~\cite{8119189, Trojannn, 8685687, 9577800, geiping2021witches, pmlr-v162-zhang22w, 9870671}. We refer to the former as \emph{deductive illusion}, reflecting that specific stimuli are derived from the victim model's general decision logic. We refer to the latter as \emph{inductive illusion}, reflecting that the victim model's general decision logic is shaped by specific stimuli. Together, these adversarial illusions represent a fundamental attack paradigm that jeopardises the trustworthiness of artificial intelligence.

Adversarial disillusion is the process of neutralising illusory stimuli, thereby minimising the risks of diverged behaviours~\cite{DBLP:conf/ndss/Xu0Q18, buckman2018thermometer, guo2018countering, samangouei2018defensegan, song2018pixeldefend, pmlr-v162-nie22a}. In essence, illusory stimuli can  be regarded as additive noise, making it natural to relate their solutions to denoising methods~\cite{10.5555/1097023, 56205, Donoho:1994aa, 1467423, 4271520, 7839189, 8579082, pmlr-v80-lehtinen18a, NEURIPS2020_4c5bcfec}. However, with the virtually limitless forms that illusory stimuli can take, it is impractical to account for every possible variation through tailored denoising methods. Addressing such multifaceted adversarial illusions calls for a unified defence framework that transcends the boundaries of individual attack forms.

In this paper, we propose a disillusion paradigm motivated by the concept of an imitation game for a universal defence against both deductive and inductive illusions. At its core lies multimodal generative agent~\cite{10.5555/3104482.3104569, JMLR:v15:srivastava14b, pmlr-v139-radford21a, pmlr-v139-jaegle21a, pmlr-v139-ramesh21a, 9878449, reed2022a}. Through the interpretative power of chain-of-thought reasoning, the multimodal generative agent is guided to observe, internalise and reconstruct the essential features of a given sample~\cite{NEURIPS2022_9d560961, NEURIPS2023_271db992, Besta_Blach_Kubicek_Gerstenberger_Podstawski_Gianinazzi_Gajda_Lehmann_Niewiadomski_Nyczyk_Hoefler_2024}. We formulate this process as an imitation game, in which the multimodal generative agent plays the role of an imitator~\cite{NEURIPS2020_1457c0d6, NEURIPS2022_b1efde53, 10.1145/3586183.3606763, Sejnowski:2023aa, Shanahan:2023aa}. While previous disillusion methods typically aim to restore the sample to its original state, prioritising perceptual similarity between the neutralised and original samples, we argue that adherence to fidelity constraint is not always indispensable and may hinder broader exploration of potential solutions. Rethinking the core principles of disillusion, we conceptualise an imitation game as an outside-the-box solution. As a proof of concept, we validate this solution with ChatGPT, a multimodal generative chatbot developed by OpenAI, through experimental simulations in the presence of various adversarial attacks.

The remainder of this paper is organised as follows. Section~\ref{sec:pre} provides a review of representative attack and defence methods, which serve as benchmarks for this study. Section~\ref{sec:method} formulates the problem and presents the proposed methodology. Section~\ref{sec:exp} evaluates the performance and discusses the experimental results. Finally, Section~\ref{sec:con} concludes the paper with a summary of the research findings and potential directions for future research.

\section{Preliminaries}\label{sec:pre}

In this section, we outline the foundational concepts that define the landscape of adversarial attacks and defences. We present a taxonomy that systematically categorises adversarial illusions based on their nature. Following this, we briefly review representative methods from both the attack and defence domains, which serve as key benchmarks in this study.

\subsection{Inference-Time Deductive Illusion}

Deductive reasoning involves forming specific conclusions from general premises. Analogously, deductive illusion entails crafting illusory stimuli derived from the victim model's decision boundary in the inference time. In this context, a malicious sample, denoted as $x'$, is generated by adding a deductive stimulus $\delta$ to a benign sample $x$, expressed as:
\begin{equation}
	x' = x + \delta .
\end{equation}
These stimuli are optimised to mislead the model's decision process while remaining imperceptible to human observers. Imperceptibility is governed by two fundamental constraints: the magnitude of distortion and the multitude of distorted components. The stimulus is generated by maximising the loss function or the distance between the model's prediction on the malicious sample and the ground truth, formulated as:
\begin{equation}
	\delta = \arg \max_{\delta} \mathcal{L}(f(x + \delta), y),
\end{equation}
subject to a magnitude constraint on the maximum allowable distortion magnitude $\epsilon_{\text{mag}}$ in the $\ell_{\infty}$-norm:
\begin{equation}
\|\delta \|_{\infty} \leq \epsilon_{\text{mag}} ,
\end{equation}
or a multitude constraint on the maximum allowable number of distorted components $\epsilon_{\text{mul}}$:
\begin{equation}
\|\delta \|_{0} \leq \epsilon_{\text{mul}}.
\end{equation}

Under the magnitude constraint, a foundational algorithm is the fast gradient sign method (FGSM)~\cite{43405}, which exploits the gradient of a loss function $\mathcal{L}$ with respect to the input sample. The stimulus is computed as:
\begin{equation}
	\delta = \epsilon_{\text{mag}} \cdot \operatorname{sign}\left(\nabla_{x} \mathcal{L}(f(x), y)\right),
\end{equation}
where $\operatorname{sign}$ represents the signum function, taking the value $+1$, $-1$ or $0$, depending on whether the gradient is positive, negative or zero. An iterative extension of this approach, known as projected gradient descent (PGD)~\cite{Madry:2018aa}, updates the stimulus at each iteration as:
\begin{equation}
	\delta = \operatorname{clip}_{\epsilon_{\text{mag}}} \left( \alpha \cdot \operatorname{sign}\left(\nabla_{x^{(t)}} \mathcal{L}(f(x^{(t)}), y)\right) \right),
\end{equation}
where $\operatorname{clip}_{\epsilon}$ denotes the clipping operator, ensuring the stimulus remains within the allowable $\ell_p$-norm ball of radius $\epsilon$, $\alpha$ represents the step size, and $x^{(t)}$ represents the input at the $t$-th iteration. Several variations of this method have been proposed. One notable strategy, termed the diverse inputs iterative fast gradient sign method (DI$^\text{2}$-FGSM)~\cite{8953423}, enhances robustness against defensive measures by applying random transformations to the input sample at each iteration:
\begin{equation}
	\delta = \operatorname{clip}_{\epsilon_{\text{mag}}} \left( \alpha \cdot \operatorname{sign}\left(\nabla_{x^{(t)}} \mathcal{L}(f(\operatorname{trans}(x^{(t)})), y)\right) \right),
\end{equation}
where $\operatorname{trans}$ represents a random transformation, such as resizing or padding. Another notable strategy, termed automatic projected gradient descent (APGD)~\cite{pmlr-v119-croce20b}, adjusts the step size adaptively to avoid extensive hyper-parameter tuning, formulated as:
\begin{equation}
	\delta = \operatorname{clip}_{\epsilon_{\text{mag}}} \left( \alpha^{(t)} \cdot \operatorname{sign}\left(\nabla_{x^{(t)}} \mathcal{L}(f(x^{(t)}), y)\right) \right),
\end{equation}
where $\alpha^{(t)}$ is the adaptive step size at the $t$-th iteration. All these attacks operate under a white-box assumption, presupposing full knowledge of the victim model, including its architecture and parameters.

Under the multitude constraint, a representative algorithm, termed OnePixel~\cite{8601309}, uses evolutionary computation to perturb the minimal number of components. At each generation, the population of candidate stimuli $\Delta$ evolves according to:
\begin{equation}
	\Delta_{\text{child}} = \operatorname{evolve}(\Delta_{\text{parent}}),
\end{equation}
where $\operatorname{evolve}$ signifies an evolutionary algorithm involving reproduction, mutation and selection rules. After the final generation, the stimulus with the highest fitness score is selected to execute the attack.

\subsection{Learning-Time Inductive Illusion}
Inductive reasoning involves forming general conclusions from specific observations. Analogously, inductive illusion entails conditioning the victim model's response with the illusory stimuli in the learning time. In this context, a malicious sample, denoted as $x'$, is created by adding a inductive stimulus $\iota$ to a benign sample $x$, expressed as:
\begin{equation}
	x' = x + \iota .
\end{equation}
A common method for reinforcing a conditioned reflex in models, known as the BadNet~\cite{8685687}, is to poison the training dataset with an arbitrary stimulus, which, when presented during inference, biases the model's predictions. Let $\mathcal{D}$ and $\mathcal{D}'$ denote the non-poisonous and poisonous sets of data, respectively. The former consists of benign samples $x$ paired with their true labels $y$, whereas the latter consists of malicious sample $x'$ paired with a target label $y'$. The model is implicitly trained to minimise a composite loss, formulated as:
\begin{equation}
	\min \sum_{(x, y) \in \mathcal{D}} \mathcal{L}(f(x), y) + \sum_{(x', y') \in \mathcal{D}'} \mathcal{L}(f(x'), y') ,
\end{equation}
where $\mathcal{L}$ is the task-specific loss function such as cross-entropy.

\subsection{Adversarial Disillusion}
Adversarial disillusion refers to the process of transforming a potentially malicious sample into a neutralised representation to mitigate the risk of biased predictions. Formally, this can be expressed as:
\begin{equation}
	\arg \min_{\tilde{x}} \| f(\tilde{x}) - f(x) \| ,
\end{equation}
where
\begin{equation}
	\tilde{x} = \operatorname{disillusion}(x') .
\end{equation}
The illusory stimuli introduced by adversarial manipulation can be regarded as a form of additive noise. Thus, it is natural to consider denoising techniques as a solution to address this threat. The simplest denoising mechanisms involve compression algorithms that remove irrelevant or less salient components, typically guided by heuristics or handcrafted parameters~\cite{guo2018countering}. More adaptive approaches utilise data-driven generative models that learn the underlying distribution and statistical characteristics of clean data through training on large datasets. Inspired by principles of thermodynamics, diffusion-based denoising models represent a class of generative methods that iteratively refine noisy samples via a diffusion process. These models employ a two-phase process comprising a forward diffusion step to deconstruct the sample and a backward diffusion step to reconstruct it. In the forward pass, Gaussian noise is incrementally added to the sample across a series of discrete steps, eventually transforming it into a distribution resembling pure Gaussian noise. During the backward pass, the model reverses this process step by step, guiding the sample back towards the distribution of clean data. Such diffusion-based mechanisms can serve as a potential solution to purify malicious samples~\cite{pmlr-v162-nie22a}.

\begin{figure*}[!t]
\centering
\includegraphics[width=0.97\linewidth]{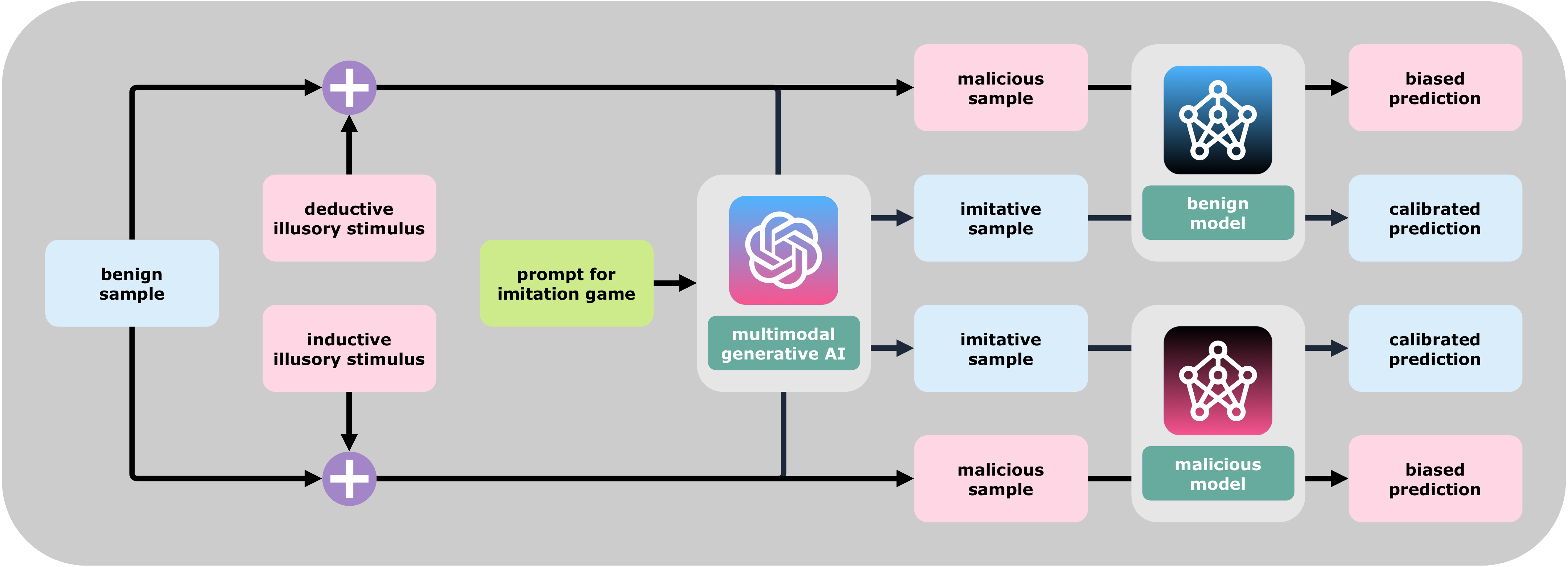}
\caption{Overview of an imitation game played by multimodal generative AI for shattering illusions induced by deductive and inductive illusory stimuli.}
\label{fig:workflow}
\end{figure*}

\section{Methodology}\label{sec:method}

Adversarial illusion attacks exploit vulnerabilities in machine learning models by embedding deductive or inductive illusory stimuli into input samples, rendering the model's predictions unreliable. To counteract these threats, we propose an imitation-based disillusion framework that leverages a multimodal generative agent that reconstructs potentially malicious samples into benign representations. We begin by formulating the problem and subsequently outline the key components of the proposed methodology.

\subsection{Problem Formulation}

Suppose a victim model is a pre-trained machine learning classifier that maps input samples to output labels, formally defined as
\begin{equation} 
f: \mathcal{X} \to \mathcal{Y} .
\end{equation}
Let $x$ denote a benign input sample from the input space $\mathcal{X}$ and $y$ the true output label from the output space $\mathcal{Y}$. An adversarially crafted sample $x'$ can be expressed as
\begin{equation}
x' = x + \delta + \iota,
\end{equation}
where $\delta$ represents a deductive stimulus and $\iota$ represents an inductive stimulus. These stimuli are designed to manipulate the victim model such that
\begin{equation}
f(x') \neq y.
\end{equation}
The objective of disillusion is to remove the illusory stimuli from $x'$, producing a neutralised sample $\tilde{x}$, ensuring that
\begin{equation}
f(\tilde{x}) = y.
\end{equation}

\subsection{Imitation Game}

Traditional disillusion methods often aim to restore the sample to its original state, prioritising perceptual similarity between the neutralised and original samples:
\begin{equation}
\tilde{x} \approx x'.
\end{equation}
While this fidelity constraint aligns with conventional notions of disillusion, we postulate that such a requirement is not invariably essential. The essence of disillusion lies not in replicating the exact semblance of the original sample but in preserving its core semantics, so that the resultant neutralised representation leads to the correct prediction. Multimodal generation opens up a promising avenue to retain the essential semantics of a sample while suppressing illusory stimuli~\cite{pmlr-v139-ramesh21a}. Let $G$ represent an arbitrary multimodal generative dialogue agent. Given an input sample $x'$ along with an instructional prompt $\pi$, the agent generates a neutralised sample $\tilde{x}$ by executing the following transformation:
\begin{equation}
\tilde{x} = G(x', \pi).
\end{equation}
Central to this transformation is the art of prompt engineering. To this end, we propose an imitation game, wherein an agent generates samples that imitate the perceived ones. To emulate a context where the objects are unknown to the agent, we leverage the role-playing capability of dialogue agents to enforce object-agnostic perception~\cite{Shanahan:2023aa}. To further elicit the agent's potential for the generative task and make the process more interpretable, we apply chain-of-thought prompting~\cite{NEURIPS2022_9d560961}, incorporating a series of intermediate reasoning steps. The prompt should instruct the agent to retain the essential features of the observed sample (e.g. a digital image) for the given task (e.g. image recognition), while eliminating malicious artefacts. A possible prompt instance is as follows:
\begin{quote}
\textit{Let's generate an imitative image step by step, assuming no prior knowledge of the main object in the provided image, as if encountering it for the first time. Begin by observing the object closely, focusing on its key features. Next, create an internal representation of the object. Finally, replicate the object in a photorealistic image, preserving its defining characteristics for accurate recognition by both humans and machines.}
\end{quote}
An overview of this imitation-based disillusion framework is illustrated in Figure~\ref{fig:workflow}.

\begin{figure*}[!t]
\centering
\includegraphics[width=0.97\linewidth]{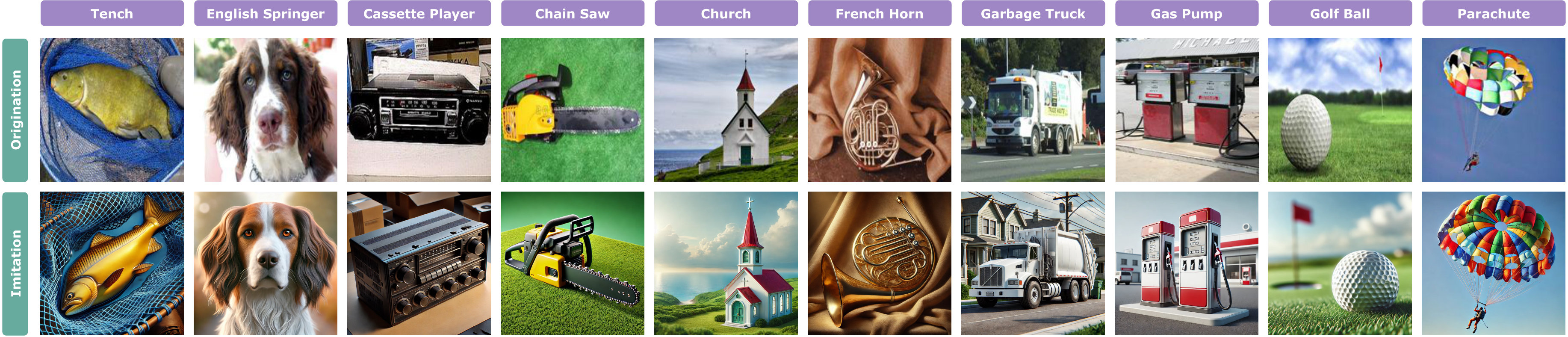}
\caption{Visual comparison between original images (top row) and imitative images (bottom row) across various object classes.}
\label{fig:exp_dataset}
\end{figure*}

\section{Experiments}\label{sec:exp}
The primary objective of the experiments is to validate the proposed imitation-based adversarial disillusion paradigm. Specifically, we aim to analyse its generalisability against different types of illusion attacks and evaluate its performance in comparison to state-of-the-art defences. We begin with a detailed experimental setup to ensure reproducibility. This is followed by a discussion of the experimental results through both visual and statistical evaluations.

\subsection{Experimental Setup}

\paragraph*{Task} Visual object recognition, a classic problem in computer vision, was selected as the fundamental task in this study. This task involves classifying objects within digital images, requiring the model under test to process and interpret visual data through accurate feature extraction and robust decision-making.

\paragraph*{Dataset} Imagenette, a curated subset of the ImageNet database~\cite{5206848}, was  chosen for the experimentation in a laboratory-scale controlled environment. It is smaller in size and scope than the full dataset to facilitate fast benchmarking and prototyping, consisting of the following 10 classes: tench, English springer, cassette player, chain saw, church, French horn, garbage truck, gas pump, golf ball, and parachute. All images were resampled to a fixed resolution of 224 × 224 pixels to standardise sample dimensions. The training set contained 9,469 samples, consistent with the standard dataset configuration, while the validation set was limited to 100 samples, with 10 samples per class, to accommodate query limits associated with the chatbot in use. The validation set was composed of correctly classified samples under the non-attack and non-defence scenario, aligning with a controlled experimental condition where the classifier's inherent errors did not bias the results of analysis.

\paragraph*{Classifier} The classification model for performing the object recognition task was the vision transformer (ViT)~\cite{Dosovitskiy:2021aa}. It is a deep learning architecture characterised by the patch-based sequential representation and self-attention mechanism. A base configuration with 16×16 pixel patches was employed for the experiments.

\paragraph*{Imitator} The imitations were performed by ChatGPT in combination with DALL·E, a multimodal generative pipeline developed by OpenAI. The vision-language prompts were interpreted via chain-of-thought reasoning~\cite{NEURIPS2022_9d560961}, and the latent representation was then used for image generation~\cite{pmlr-v139-ramesh21a}.

\begin{figure*}[!t]
\centering
\includegraphics[width=0.98\linewidth]{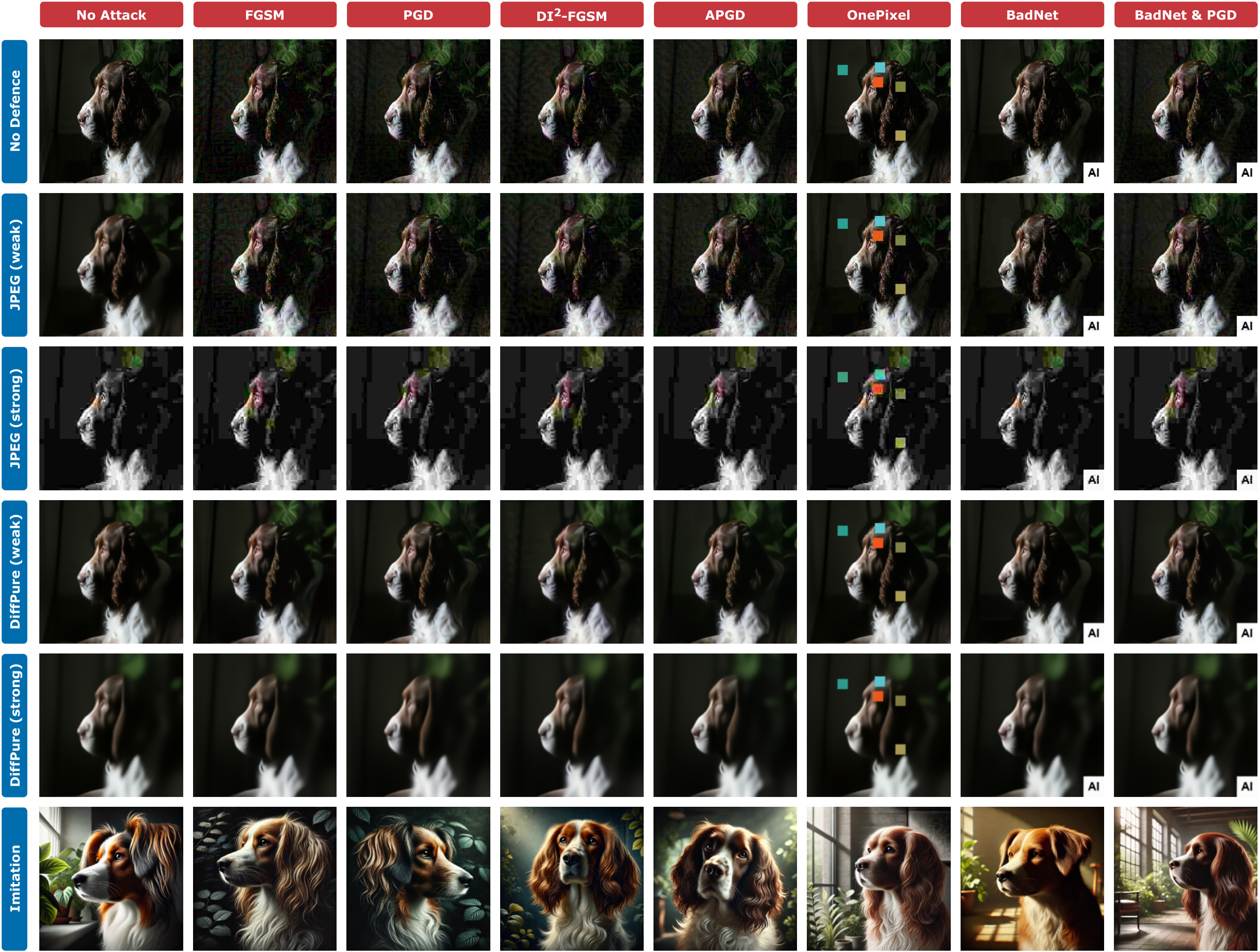}
\caption{Visual comparison of multiple defence methods (rows) against multiple attack methods (columns).}
\label{fig:exp_vis}
\end{figure*}

\paragraph*{Attacks} The attacks were simulated on two variants of the classifier: a benign model (without a backdoor) and a malicious model (with a backdoor). Non-targeted attacks, which aim to cause misclassification without targeting a specific incorrect class, were applied to the benign model. These included both white-box magnitude-limited and black-box multitude-limited deductive attacks. The white-box attacks comprised FGSM~\cite{43405}, PGD~\cite{Madry:2018aa}, DI$^{\text{2}}$-FGSM~\cite{8953423} and APGD~\cite{pmlr-v119-croce20b}, with the magnitude limit $\epsilon_{\text{mag}}$ set to 8/255 for all pixels. The black-box attack was represented by the OnePixel method~\cite{8601309}, with the multitude limit $\epsilon_{\text{mul}}$ set to 5 patches of 15 × 15 pixels. Targeted attacks, which are designed to mislead the classifier into predicting a specific incorrect class, were applied to the malicious model. These included an inductive attack using BadNet~\cite{8685687} and a hybrid attack implemented by combining BadNet with PGD. The BadNet poisoning rate, defined as the percentage of malicious training data, was set to 10\%, and the triggering stimulus was configured as an all-white pattern of 32 × 32 pixels with the text `AI' at the centre, positioned in the bottom-right corner of each image sample.

\paragraph*{Defences} The defence methods under comparison comprised a heuristic-based approach and a data-driven generative model. The heuristic-based approach was exemplified by JPEG compression~\cite{guo2018countering}, while the data-driven model was represented by DiffPure~\cite{pmlr-v162-nie22a}. Each defence method was implemented with both weak and strong variants. For JPEG, the quality factors were set to 25 for the weak variant and 5 for the strong variant. For DiffPure, the Gaussian standard deviations in the diffusion schedule were configured to 0.1 for the weak variant and 0.3 for the strong variant.

\paragraph*{Metrics} The overall performance of individual defence methods against various attack scenarios was evaluated based on classification accuracy. For the benign model under non-targeted attacks, accuracy was defined as the ratio of correctly classified samples to the total number of samples:
    \begin{equation}
    \text{ACC}_{\text{non-target}} = \frac{\text{correct classifications}}{\text{all classifications}}.
    \end{equation}
	For the malicious model under targeted attacks, accuracy was defined as the ratio of correctly classified samples to the total number of samples, excluding those belonging to the attack target class:
    \begin{equation}
    \text{ACC}_{\text{target}} = \frac{\text{correct classifications}}{\text{all classifications except attack target}}.
    \end{equation}

\subsection{Experimental Results}
\paragraph*{Visual Analysis}
Figure~\ref{fig:exp_dataset} illustrates a comparison between original and imitative images across various object classes. The imitative images were generated by the multimodal generative agent through a step-by-step replication process that captures the defining characteristics of the original objects. These images aim to preserve recognisability while being computationally synthesised, visually demonstrating the generative model's ability to depict key features of the objects, such as texture, shape, colour and context. Figure~\ref{fig:exp_vis} provides a comparative visual analysis of the resulting images produced by applying specific defences against particular attacks, demonstrating the interaction between attack and defence mechanisms. Different attacks exhibit varying noise and corruption patterns, reflecting their distinct illusory mechanisms. Strong variants of the defences demonstrated a more pronounced noise removal effect compared to their weaker counterparts. However, this came at the cost of a deterioration in recognition features as a trade-off. In contrast, the images generated via imitation, rather than adhering to a strict spatial fidelity constraint, preserved the primary features while leaving behind negligible, if any, adversarial noise.

\paragraph*{Statistical Analysis} 

\begin{figure*}[!t]
\centering
\includegraphics[width=0.92\linewidth]{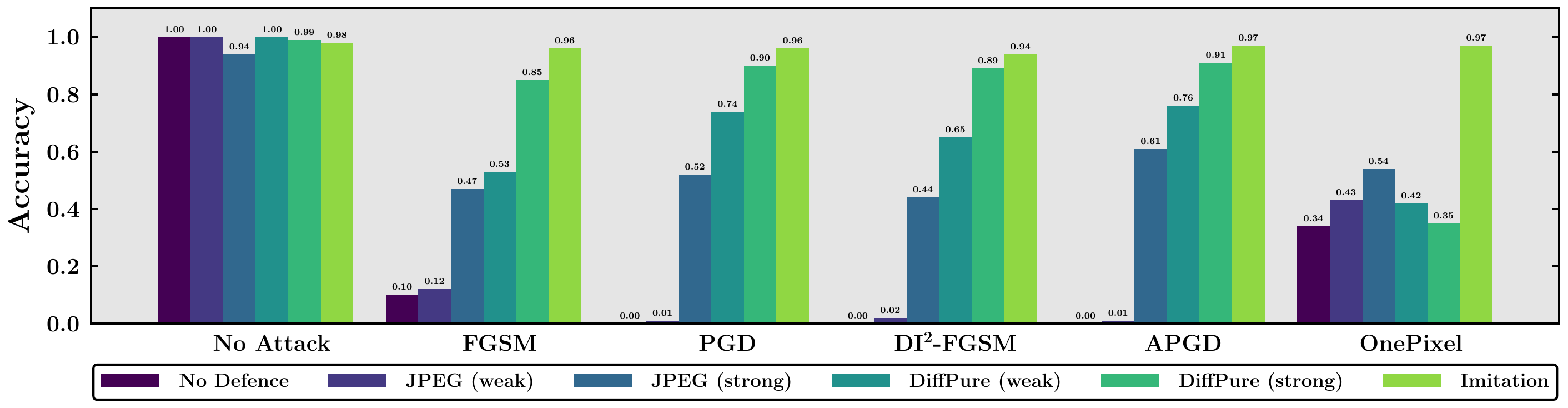}
\caption{Accuracy of the benign classifier under various non-targeted attack methods, evaluated with various defence methods.}
\label{fig:exp_nontarget}
\end{figure*}

\begin{figure*}[!t]
\centering
\includegraphics[width=0.92\linewidth]{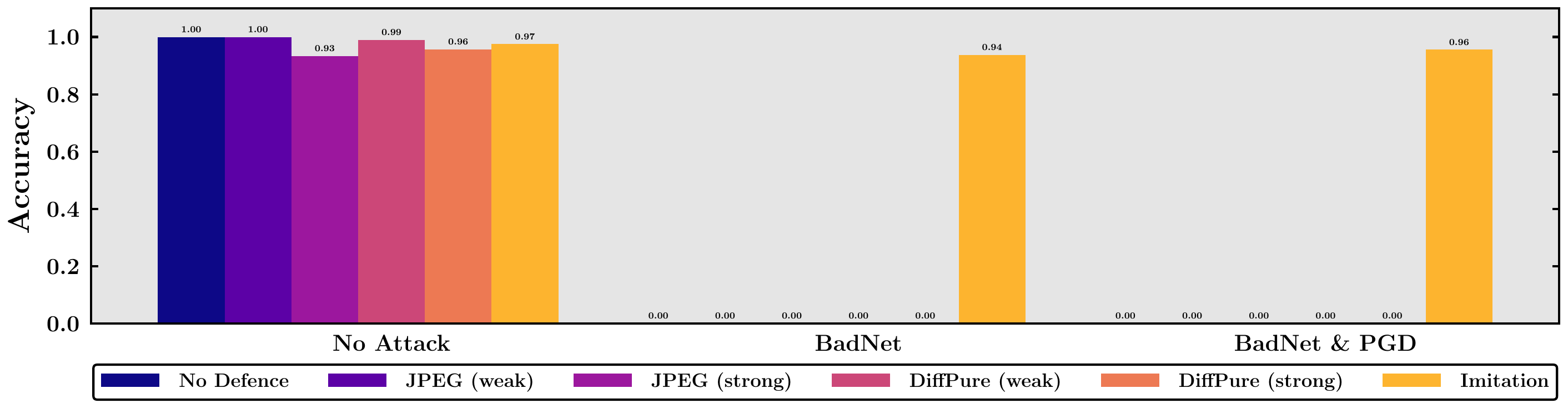}
\caption{Accuracy of the malicious classifier under various targeted attack methods, evaluated with various defence methods.}
\label{fig:exp_target}
\end{figure*}

The performance of various defence methods was evaluated in terms of classification accuracy under multiple adversarial attack scenarios. As illustrated in Figure~\ref{fig:exp_nontarget} and Figure~\ref{fig:exp_target}, experiments were conducted on two models: a benign classifier subjected to non-targeted attacks and a malicious classifier exposed to targeted attacks.

Without any defence, both classifiers maintained perfect accuracy of 100\% in the absence of attacks. However, once adversarial stimuli were introduced, the accuracy dropped drastically, ranging from 0\% to approximately 30\%. These results reflect the vulnerability of machine perception to illusory stimuli, which can effectively redirect model predictions.

Benchmark defensive mechanisms, including JPEG compression and DiffPure, provided partial mitigation. In most cases, the strong variants of each method outperformed their weaker counterparts. For non-targeted white-box deductive illusions (FGSM, PGD, DI$^\text{2}$-FGSM, and APGD), JPEG (strong) increased accuracy to a range of approximately 40\% to 60\%, while DiffPure (strong) further improved it to a range of approximately 85\% to 90\%. However, this improvement came at the cost of a slight reduction in accuracy under no-attack conditions, indicating a trade-off between robustness and fidelity. For the non-targeted black-box deductive illusion (OnePixel), the attack success rate was lower compared to the white-box attacks, with accuracy dropping to 34\% without any defence. However, it remained difficult to counteract, as the accuracy recovery achieved by JPEG and DiffPure was limited, yielding values only in the range of 35\% to 54\%. In the case of targeted attacks, including the inductive illusion (BadNet) and the hybrid illusion (BadNet \& PGD), both defensive mechanisms failed to correct the predictions, as accuracy remained at 0\%, consistent with the scenario where no defence was applied.

Among all evaluated defences, the imitation-based method consistently delivered the most robust performance. On the benign classifier, it achieved 94\% to 97\% accuracy across all non-targeted attacks. On the malicious classifier, it successfully restored 94\% and 96\% accuracy under the targeted attacks, far surpassing all benchmark methods. It also maintained near-perfect accuracy under no-attack conditions, demonstrating minimal degradation when no threat was present. These results confirm that the imitation-based disillusion framework effectively neutralises both deductive and inductive adversarial illusions, thereby preserving the trustworthiness of machine perception. Its consistent effectiveness across diverse attack types, coupled with low collateral impact, validates its promise as a general-purpose defence strategy.

\section{Conclusions}\label{sec:con}
In this study, we introduced an adversarial disillusion paradigm based on the imitation ability of multimodal generative artificial intelligence. Experimental results validated its generalisability against various types of illusion attacks and demonstrated a notable improvement over prior art. Visual analysis further confirmed that our method effectively preserved essential semantic features while suppressing adversarial artefacts, without relying on strict spatial fidelity. A limitation of this study is the difficulty in perfectly emulating an environment where a multimodal generative agent has no prior knowledge of an object. Although an attempt to impose this assumption has been made through prompting, it remains challenging to analyse the agent's imitation ability in the presence of completely unknown objects. This difficulty arises due to the vastness of the agent's learning set, which, while generally advantageous in practical real-world applications, complicates controlled analysis. We further note that, at the time of writing, regulations on the use of generative artificial intelligence, driven by ethical concerns surrounding disinformation, could constrain the agent's potential ability to perform imitation. Future research may assess whether sophisticated attacks might penetrate the disillusion mechanism offered by the generative agent, potentially compromising the security of the protected machine learning models. We envision that the concept of the imitation game will pave the way for broadening the scope of adversarial disillusion, advancing the quest for a more universal solution capable of countering the myriad adversarial attacks faced in dynamic real-world environments.

\section*{Acknowledgements}
This work was supported in part by the Japan Society for the Promotion of Science (JSPS) under KAKENHI Grants (JP21H04907 and JP24H00732), and in part by the Japan Science and Technology Agency (JST) under CREST Grant (JPMJCR20D3) including AIP Challenge, AIP Acceleration Grant (JPMJCR24U3) and K Program Grant (JPMJKP24C2).

\bibliography{Transactions-Bibliography/bstcontrol, Bib/bib_imit}
\bibliographystyle{Transactions-Bibliography/IEEEtran}

\vspace{-10em}
\begin{IEEEbiographynophoto}{Ching-Chun Chang} received the PhD in Computer Science from the University of Warwick, UK, in 2019. He is currently affiliated with the National Institute of Informatics, Japan, as a Project Assistant Professor. He participated in the Short-Term Scientific Mission supported by European Cooperation in Science and Technology Actions at the Faculty of Computer Science, Otto von Guericke University of Magdeburg, Germany, in 2016. He was granted the Marie-Curie Fellowship and participated in the Research and Innovation Staff Exchange supported by Marie Skłodowska-Curie Actions at the Department of Electrical and Computer Engineering, New Jersey Institute of Technology, USA, in 2017. He was a Visiting Scholar at the School of Computing and Mathematics, Charles Sturt University, Australia, in 2018, and at the School of Information Technology, Deakin University, Australia, in 2019. He was a Research Fellow at the Department of Electronic Engineering, Tsinghua University, China, in 2020. His research interests include artificial intelligence, biometrics, cryptography, cybersecurity, evolutionary computation, forensics, information theory, steganography, and watermarking.
\end{IEEEbiographynophoto}

\newpage
\begin{IEEEbiographynophoto}{Fan-Yun Chen} 
received the BS degree from the Department of Applied Mathematics, National Chiayi University, Taiwan, in 2023. He is currently pursuing the MS degree in the Department of Information Engineering and Computer Science, Feng Chia University, Taiwan. His research interests include artificial intelligence, cybersecurity, and adversarial machine learning.
\end{IEEEbiographynophoto}

\begin{IEEEbiographynophoto}{Shih-Hung Ku} 
received the BS degree from the Department of Applied Mathematics, Feng Chia University, Taiwan, in 2024. He is currently pursuing the MS degree in the Department of Information Engineering and Computer Science, Feng Chia University, Taichung, Taiwan. His research interests include artificial intelligence, cybersecurity, and adversarial machine learning.
\end{IEEEbiographynophoto}

\begin{IEEEbiographynophoto}{Kai Gao} 
received the BSc degree in Software Engineering from Fujian Normal University, China, in 2018. He is currently pursuing the PhD degree in the Department of Information Engineering and Computer Science, Feng Chia University. His research interests include cryptography, cybersecurity, secret sharing, steganography, and machine learning.
\end{IEEEbiographynophoto}

\begin{IEEEbiographynophoto}{Hanrui Wang}
received the BS degree in Electronic Information Engineering from Northeastern University, China, in 2011. He left the IT industry from a director position in 2019 to pursue a research career and received the PhD in Computer Science from Monash University, Australia, in January 2024. He is currently working as a Postdoctoral Researcher at the National Institute of Informatics, Tokyo, Japan. His research interests include AI security and privacy, particularly adversarial machine learning.
\end{IEEEbiographynophoto}

\begin{IEEEbiographynophoto}{Isao Echizen} received BS, MS, and DE degrees from the Tokyo Institute of Technology, Japan, in 1995, 1997 and 2003, respectively. He joined Hitachi, Ltd. in 1997 and until 2007 was a Research Engineer in the company's systems development laboratory. He is currently a Director and Professor of the Information and Society Research Division, as well as a Director of the Global Research Center for Synthetic Media, at the National Institute of Informatics; a Professor in the Department of Information and Communication Engineering, Graduate School of Information Science and Technology, the University of Tokyo; and a Professor in the Graduate University for Advanced Studies (SOKENDAI), Japan. He was a Visiting Professor at the Tsuda University, Japan; at the University of Freiburg, Germany; and at the University of Halle-Wittenberg, Germany. He is currently engaged in research on AI security, multimedia security and multimedia forensics, serving as a Research Director for the CREST FakeMedia project and the K Program SYNTHETIQ X project of the Japan Science and Technology Agency (JST). He received the Commendation for Science and Technology by the Minister of Education, Culture, Sports, Science and Technology (Research Category) in 2025. He also received the IEICE Best Paper Award in 2023; the IPSJ Best Paper Awards in 2005 and 2014; the IPSJ Nagao Special Researcher Award in 2011; the DOCOMO Mobile Science Award in 2014; the IISEC Information Security Cultural Award in 2016; and the IEEE WIFS Best Paper Award in 2017. He is an IEICE Fellow, an IPSJ Fellow, an IEEE Senior Member, an IFIP Japanese Representative, and an APSIPA Vice President.
\end{IEEEbiographynophoto}

\end{document}